\documentclass[runningheads]{llncs}
\usepackage[latin9]{inputenc}
\usepackage{booktabs}
\usepackage{url}
\usepackage{graphicx}

\makeatletter

\DeclareTextSymbolDefault{\textquotedbl}{T1}
\providecommand{\tabularnewline}{\\}
\newcommand{\lyxdot}{.}

\usepackage[english]{babel}

\@ifundefined{showcaptionsetup}{}{%
 \PassOptionsToPackage{caption=false}{subfig}}
\usepackage{subfig}
\makeatother

\begin{document}
%===========================================================

\title{Totally Looks Like - How Humans Compare, Compared to Machines\thanks{This research was supported through grants to the senior author, for
which all authors are grateful: Air Force Office of Scientific Research
(FA9550-18-1-0054), the Canada Research Chairs Program (950-219525),
the Natural Sciences and Engineering Research Council of Canada (RGPIN-2016-05352)
and the NSERC Canadian Network on Field Robotics (NETGP417354-11).}}

\author{Amir Rosenfeld \and Markus D. Solbach \and John K. Tsotsos\\{\texttt {\{amir, solbach, tsotsos\}@cse.yorku.ca}}}
\institute{Department of Electrical Engineering and Computer Science}
\institute{York University\\
Toronto, ON, Canada, M3J 1P3}

\maketitle
% Replace your paper's title here

\titlerunning{Short paper title}

% Replace an abstracted version of your paper's title here

%===========================================================

%Please include author names in full in the paper, 
%If any authors have names that can be parsed into FirstName LastName in multiple ways, please include the correct parsing, in a comment to the volume editors:
%\index{Lastnames, Firstnames}

\authorrunning{A. Rosenfeld, M. Solbach et al.}

% A shorter version of authors' name
% First names are abbreviated in the running head.
% If there are more than two authors, 'et al.' is used.

%===========================================================

%===========================================================

\begin{abstract}
Perceptual judgment of image similarity by humans relies on rich internal
representations ranging from low-level features to high-level concepts,
scene properties and even cultural associations. However, existing
methods and datasets attempting to explain perceived similarity use
stimuli which arguably do not cover the full breadth of factors that
affect human similarity judgments, even those geared toward this goal.
We introduce a new dataset dubbed \textbf{Totally-Looks-Like} (TLL)
after a popular entertainment website, which contains images paired
by humans as being visually similar. The dataset contains 6016 image-pairs
from the wild, shedding light upon a rich and diverse set of criteria
employed by human beings. We conduct experiments to try to reproduce
the pairings via features extracted from state-of-the-art deep convolutional
neural networks, as well as additional human experiments to verify
the consistency of the collected data. Though we create conditions
to artificially make the matching task increasingly easier, we show
that machine-extracted representations perform very poorly in terms
of reproducing the matching selected by humans. We discuss and analyze
these results, suggesting future directions for improvement of learned
image representations. 
\end{abstract}

%%%%%%%%% BODY TEXT

\section{Introduction}
\begin{center}
\begin{figure*}
\caption{\label{fig:The-Totally-Looks-Like-dataset}The \emph{Totally-Looks-Like
}dataset: pairs of perceptually similar images selected by human users.
The pairings shed light on the rich set of features humans use to
judge similarity. Examples include (but are not limited to): attribution
of facial features to objects and animals \emph{(a,b)}, global shape
similarity \emph{(c,d)}, near-duplicates \emph{(e)}, similar faces
\emph{(f)}, textural similarity \emph{(g), }color similarity\emph{
(h)}}

\begin{centering}
\begin{minipage}[t]{0.24\textwidth}%
\includegraphics[viewport=0bp 23bp 401bp 271bp,clip,scale=0.25]{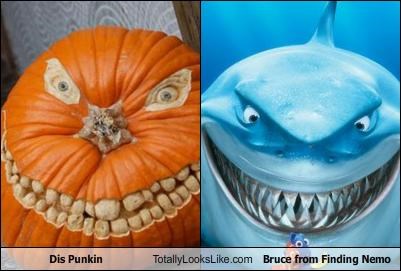} (a) 
\end{minipage}\,%
\begin{minipage}[t]{0.24\textwidth}%
\includegraphics[viewport=0bp 23bp 401bp 271bp,clip,scale=0.25]{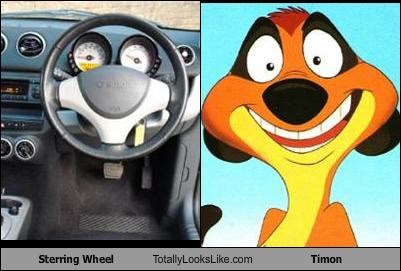} 
(b) 
\end{minipage}\,%
\begin{minipage}[t]{0.24\textwidth}%
\includegraphics[viewport=0bp 23bp 401bp 271bp,clip,scale=0.25]{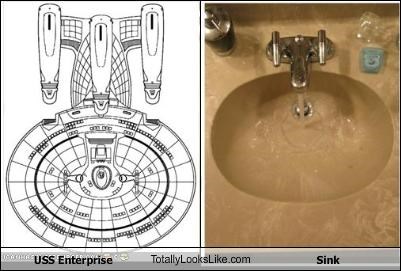} 
(c) 
\end{minipage}\,%
\begin{minipage}[t]{0.24\textwidth}%
\includegraphics[viewport=0bp 23bp 401bp 271bp,clip,scale=0.25]{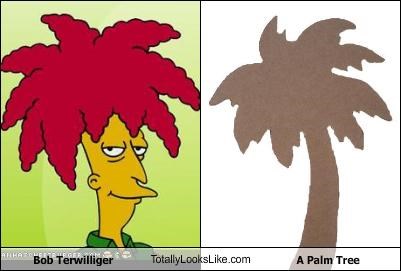} 
(d) 
\end{minipage}
\par\end{centering}
\centering{}%
\begin{minipage}[t]{0.24\textwidth}%
\includegraphics[viewport=0bp 23bp 401bp 271bp,clip,scale=0.25]{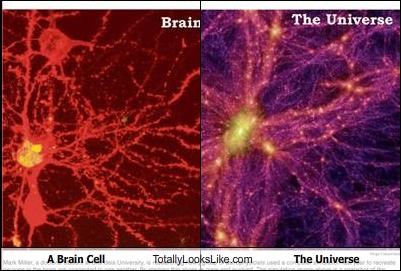} 
(e) 
\end{minipage}\,%
\begin{minipage}[t]{0.24\textwidth}%
\includegraphics[viewport=0bp 23bp 401bp 271bp,clip,scale=0.25]{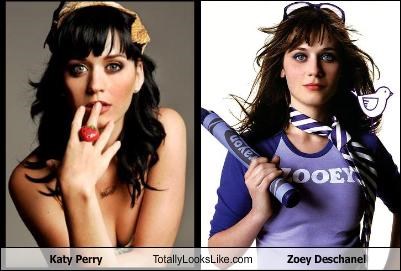} 
(f) 
\end{minipage}\,%
\begin{minipage}[t]{0.24\textwidth}%
\includegraphics[viewport=0bp 23bp 401bp 271bp,clip,scale=0.25]{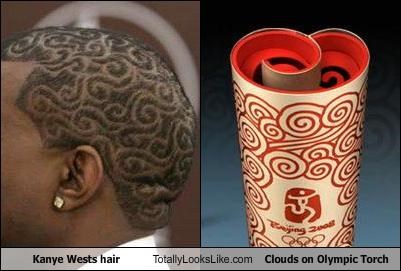} 
(g) 
\end{minipage}\,%
\begin{minipage}[t]{0.24\textwidth}%
\includegraphics[viewport=0bp 23bp 401bp 271bp,clip,scale=0.25]{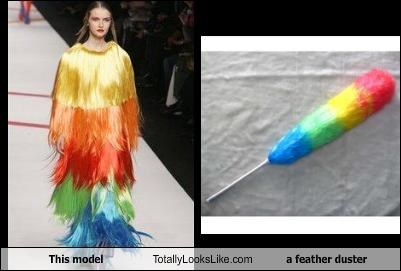} 

(h) 

\end{minipage}
\end{figure*}
\par\end{center}
Human perception of images goes far beyond objects, shapes, textures
and contours. Viewing a scene often elicits recollection of other
scenes whose global properties or relations resemble the currently
observed one. This relies on a rich representation of image space
in the brain, entailing scene structure and semantics, as well as
a mechanism to use the representation of an observed scene to recollect
similar ones from the profusion of those stored in memory. Though
not fully understood, the capacity of the human brain to memorize
images is surprisingly large \cite{brady2008visual,konkle2010scene}.
The recent explosion in the performance and applicability of deep-learning
models in all fields of computer vision \cite{schmidhuber2015deep,zhou2017landscape,liu2017survey}
(and others), including image retrieval and comparison \cite{zhou2017recent},
can tempt one to conclude that the representational power of such
methods approaches that of humans, or perhaps even exceeds them. We
aim to explore this by testing how deep neural networks fare on the
challenge of similarity judgment between pairs of images from a new
dataset, dubbed \textquotedbl\textbf{Totally-Looks-Like}\textquotedbl{}
(TLL); See Figure \ref{fig:The-Totally-Looks-Like-dataset}. It is
based on a website for entertainment purposes, which hosts pairs of
images deemed by users to appear similar to each other, though they
often share little common appearance, if judging by low-level visual
features. These include pairs of images out of (but not limited to)
objects, scenes, patterns, animals, and faces across various modalities
(sketch, cartoon, natural images). The website also includes user
ratings, showing the level of agreement with the proposed resemblances.
Though it is not very large, the diversity and complexity of the images
in the dataset implicitly captures many aspects of human perception
of image similarity, beyond current datasets which are larger but
at the same time narrower in scope. We evaluate the performance of
several state-of-the-art models on this dataset, cast as a task of
image retrieval. We compare this with human similarity judgments,
forming not only a baseline for future evaluations, but also revealing
specific weaknesses in the strongest of the current learned representations
that point the way for future research and improvements. We conduct
human experiments to validate the consistency of the collected data.
Even though in some experiments we allow very favorable conditions
for the machine-learned representations, they still often fall short
of correctly predicting the human matches.

The next section overviews related work. This is followed by a description
of our method, experiments and analysis. We close the paper with discussion
about the large gaps between what is expected of state-of-the art
learned representations and suggestions for future work. The dataset
is available at the following address: \url{https://sites.google.com/view/totally-looks-like-dataset}

\section{Related Work}

This paper belongs to a line of work that compares machine and human
vision (in the context of perception) or attempts to perform some
vision related task that is associated with high-level image attributes.
As ourselves, others also tapped the resources of social media/online
entertainment websites to advance research in high-level image understanding.
For example, Deza and Parikh \cite{deza2015understanding} collected
datasets from the web in order to predict the virality of images,
reporting super-human capabilities when five high-level features were
used to train an SVM classifier to predict virality.

Several lines of work measure and analyze differences between human
and machine perception. The work of \cite{pramod2016computational}
collected 26k perceived dissimilarity measurements from 2,801 visual
objects across 269 human subjects. They found several discrepancies
between computational models and human similarity measurements. The
work of \cite{10.3389/fpsyg.2017.01726} suggests that much of human-perceived
similarity can readily be accounted for by representations emerging
in deep-learned models. Others modify learned representations to better
match this similarity, reporting a high-level of success in some cases
\cite{peterson2016adapting}, and near-perfect in others \cite{battleday2017modeling}.
The work of \cite{battleday2017modeling} is done in a context which
reduces similarity to categorization. Very recently, Zhang Et al.
\cite{zhang2018unreasonable} have shown that estimation of human
perceptual similarity is dramatically better using deep-learned features,
whether they are learned in a supervised or unsupervised manner, than
more traditional methods. Their evaluation involved comparing images
to their distorted versions. The distortions tested were quite complex
and diverse. Akin to ours, there are works who question the behavioral
level of humans vs. machines. For instance, Das et. al \cite{das2017human}
compare the attended image regions in Visual Question Answering (VQA,
\cite{antol2015vqa}) to that of humans and report a rather low correlation.
Other works tackle high level tasks such as understanding image aesthetics
\cite{workman2016quantifying} or even humor \cite{chandrasekaran2016we}.
The authors of \cite{geirhos2017comparing} compare the robustness
of humans vs. machines to image degradations, showing that DNN's that
are not trained on noisy data are more error-prone than humans, as
well as having a very different distribution of non-class predictions
when confronted with noisy images. Matching images and recalling them
are two very related subjects, as it seems unlikely for a human (or
any other system storing a non-trivial amount of images) to perform
exhaustive search over the entire collection of images stored in memory.
Studies of image memorability \cite{ICCV15_Khosla} have successfully
produced computational models to predict which images are more memorable
than others.

The works of \cite{pramod2016computational,peterson2016adapting,10.3389/fpsyg.2017.01726,zhang2018unreasonable}
show systematic results on large amounts of data. However, most of
the images within them either involve objects with a blank background
\cite{pramod2016computational,10.3389/fpsyg.2017.01726} or of a narrow
type (e.g., animals \cite{peterson2016adapting}). Our dataset is
smaller in scale than most of them, but it features images from the
``wild'', requiring similarities to be explained by features ranging
from low-level to abstract scene properties. In \cite{zhang2018unreasonable},
a diverse set of distortions is applied to images, however, the source
image always remains the same, whereas the proposed dataset shows
pairs of images of different scenes and objects, still deemed similar
by human observers. In this context, the proposed dataset does not
contradict the systematic evaluations performed by prior art, but
rather complements them and broadens the scope to see where modern
image representations still fall short.

\section{Method}

The main source of data for the reported experiments is a popular
website called \emph{TotallyLooksLike}\footnote{\url{http://memebase.cheezburger.com/totallylookslike}}.
The website describes itself simply as ``Stuff That Looks Like Other
Stuff''. For the purpose of amusement, users can upload pairs of
images which, in their judgment, resemble each other. Such images
may be have any content, such as company logos, household objects,
art-drawing, faces of celebrities and others. Figure \ref{fig:The-Totally-Looks-Like-dataset}
shows a few examples of such image pairings. Each submission is shown
on the website, and viewers can express their agreement (or disagreement)
about the pairing by choosing to up-vote or down-vote. The total number
of up-votes and down-votes for each pair of images is displayed.

Little do most of the casual visitors of this humorous website realize
that it is in fact a hidden treasure: humans encounter an image in
the wild and recall another image which not only do they deem similar,
but so do hundreds of other site users (according to the votes). This
provides a dataset of thousands of such image pairings, by definition
collected from the wild, that may aid to explore the cognitive drive
behind judgment of image similarity. Beyond this, it contains samples
of images that one recollects when encountering others, allowing exploration
in the context of long-term visual memory and retrieval.

While other works have explored image memorability \cite{ICCV15_Khosla},
in this work we focus on the aspects of similarity judgment. We next
describe the dataset we created from this website.

\subsection{Dataset}

We introduce the \emph{Totally-Looks-Like }(\textbf{TLL}) dataset.
The dataset contains a snapshot of 6016 image-pairs along with their
votes downloaded from the website in Jan. 2018 (a few images are added
each day). The data has been downloaded with permission from the web-site's
administrators to make it publicly available for research purposes.
For each image pair, we simply refer to the two images as the ``left
image'' and the ``right image'', or more concisely as $<L_{i},R_{i}>,i\in1\dots N$
where N is the total number of images in the dataset. We plan to make
the data available on the project website, along with pre-computed
features which will be listed below.

\subsection{Image Retrieval}

The TLL dataset is the basis for our experiments. We wish to test
to what degree similarity metrics based on generic machine-learned
representations are able to reproduce the human-generated pairings.

We formulate this as a task of image retrieval: Let $\mathcal{L}=(L_{i})_{i}$
be the set of all left images and similarly let $\mathcal{R}$ be
the set of all right images. For a given image $L_{i}$ we measure
the distance $\phi(L_{i},R_{j})$ between $L_{i}$ and each $R_{j}\in\mathcal{R}$.
This induces a ranking $r_{1},\dots r_{n}$ over $\mathcal{R}$ by
sorting according to the distance $\phi(\cdot,\cdot)$. A perfect
ranking returns $r_{1}=i$. Calculating distances using $\phi$ over
all pairs of the dataset allows us to measure its overall performance
as a distance metric for retrieval. For imperfect rankings, we can
measure the recall up to some ranking $k$, which is the average number of
times the correct match was in the top-$k$ ranked images. In practice,
we measure distances between feature representations extracted via
state-of-the-art DCNN's, either specialized for generic image categorization
or face identification, as detailed in the experiments section.

\textbf{Direct Comparison vs. Recollection\label{par:Direct-Comparison-vs.}}:
We note that framing the task as image retrieval may be unfair to
both sides: when humans encounter an image and recollect a perceptually
similar one to post on the website, they are not faced with a forced
choice task of selecting the best match out of a predetermined set.
Instead, the image triggers a recollection of another image in their
memory, which leads to uploading the image pair. On one hand, this
means that the set of images from which a human selects a match is
dramatically larger than the limited-size dataset we propose, so the
human can potentially find a better match. On the other hand, the
human does not get to scrutinize each image in memory, as the process
of recollection likely happens in an associative manner, rather than
by performing an exhaustive search on all images in memory. In this
regard, the machine is more free to spend as many computational resources
as needed to determine the similarity between a putative match. Another
advantage for the machine is that the ``correct'' match already
exists in the predetermined dataset; possibly finding it will be easier
than in an open-ended manner as a human does. Nevertheless, we view the task of retrieval from this closed set as a first approximation. In addition, we suggest below some ways to
make the comparison more fair.

\section{Experiments}

We now describe in detail our experiments, starting from data collection
and preprocessing, through various attempts to reproduce the human
data and accompanying analysis.

\textbf{Data Preprocessing} All images if the TLL (Totally-Looks-Like)
dataset were automatically downloaded along with their up-votes and
down-votes from the website. Each image pair $<L_{i},R_{i}>$appears
on the website as a single image showing $L_{i}$ and $R_{i}$ horizontally
concatenated, of constant width of 401 pixels and height of 271 pixels.
We discard for each image the last column and split it equally to
left and right images. In addition, the bottom 26 pixels of each image
contains for each side a description of the content. While none of
the methods we apply explicitly use any kind of text detection/recognition,
we discard these rows as well to avoid the possibility of ``cheating''
in the matching process.

\subsection{Feature extraction}

We extract two kinds of features from each image: generic and facial.

\textbf{Generic Features}: we extract ``generic'' image features
by recording the output of the penultimate layer of various state-of-the-art
network architectures for image categorization, trained on the ImageNet
benchmark \cite{russakovsky2015imagenet}, which contains more than
a million training images spread over a thousand object categories.
Training on such a rich supervised task been shown many times to produce
features which are transferable across many tasks involving natural
images \cite{sharif2014cnn}. Specifically, we use various forms of
Residual Networks \cite{he2016deep}, Dense Residual Networks,\cite{huang2016densely},
AlexNet \cite{krizhevsky2012imagenet} and VGG-16 from \cite{simonyan2014very},
giving rise to feature-vector dimensionalities ranging from a few
hundred to a few thousands, dependent on the network architecture.
We extract the activations of the penultimate layer of each of these
networks for each of the images and store them for distance computations.

\textbf{Facial Features}: many of the images contain faces, or objects
that resemble faces. Faces play an important role in human perception
and give rise to many of the perceived similarities. We run a face
detector on all images, recording the location of the face. For each
detected face in each image, we extract features using a deep neural
network which was specifically designed for face recognition. The
detector and features both use an off-the-shelf implementation \footnote{\url{https://github.com/ageitgey/face_recognition}}.
The dimensionality the extracted face descriptor is 128. Figure \ref{fig:Probability-of-agreement}
(c) shows the distribution of the number of detected faces in images,
as well as the agreement between the number of detected faces in human-matched
pairs. The majority of images have a face detected in them, which
very few containing more than one face. When a face is detected in
a left image of a given pair, it is likely that a face will be detected
in the right one as well.

\textbf{Generic-Facial Features} : very often in the TLL dataset,
we can find objects that resemble faces and play an important role
in these images, being the main object which led to the selection
of an image pair. To allow comparing such objects to one another,
we extract generic image features from them, as described above, to
complement the description by specifically tailored facial features.
We do this under the likely assumption that while a facial feature
extractor might not produce reliable features for comparison from
a face-like object (because the network was not trained on such images),
a generic feature extractor might.

We denote by $G_{i}$, $F_{i}$, and $GF_{i}$ the set of generic
features, facial features and generic-facial features extracted from
each image. Note that for some images faces are not at all detected,
and so $F_{i}$ and $GF_{i}$ are empty sets. For others, possibly
more than one face is detected, in which $F_{i}$ and $GF_{i}$ can
be sets of features.

We next describe how we take all of these features into account.

\subsection{Matching Images}

We define the distance function between a pair of images $L_{i}$,
$R_{j}$ by their extracted features as described above. We either
use the $\ell_{2}$ (Euclidean) distance between a pair of features,
i.e., $\phi_{l}^{f}(A,B)=\|A-B\|_{2}$ or the cosine distance, i.e,
$\phi_{c}^{f}(A,B)=1-\frac{A\cdot B}{\|A\|\|B\|}$. Where $A,B$ are
the corresponding features for images $L_{i}$, $R_{j}$. The subscripts
$l,c$ specify $\ell_{2}$ norm or cosine distance. The superscript
$f$ specifies the kind of representation used, i.e, $f\in\{G,F,GF\}$.
For facial features ($F$) we use only the euclidean distance, as
is designated by the applied facial recognition method. Each distance
function $\phi_{l}^{f}$ generates a distance matrix $\Phi_{l}^{f}\in\mathcal{\mathcal{R}}^{N}$
with the $i,j$ location representing the distance between $L_{i}$,
$R_{j}$ using this function. For image pairs with more than one face
in either image we assign the corresponding $i,j$ location the minimal
distance between all pairs of features extracted from the corresponding
faces. For image pairs where at least one image has no detected face
we assign the corresponding distance to $+\infty$.

Armed with $\Phi_{l}^{f}$, we may now test how the distance-induced
ranking aligns with the human-selected matches.

\begin{figure}
\caption{\label{fig:cosine vs l2}(a) Difference between recall per number
of images retrieved for cosine and $\ell_{2}$-distance based retrieval.
Recall is always improved if we use the cosine distance over the $\ell_{2}$
distance between representations. \label{fig:Retrieval-Graph}(b)
Retrieval performance by various learned representations in the TLL
dataset. Left: all images. Right: showing recall only for the top
1 (first place), 5, 10, 20 images. }

\centering{}\subfloat[]{\begin{centering}
\includegraphics[height=0.28\textwidth]{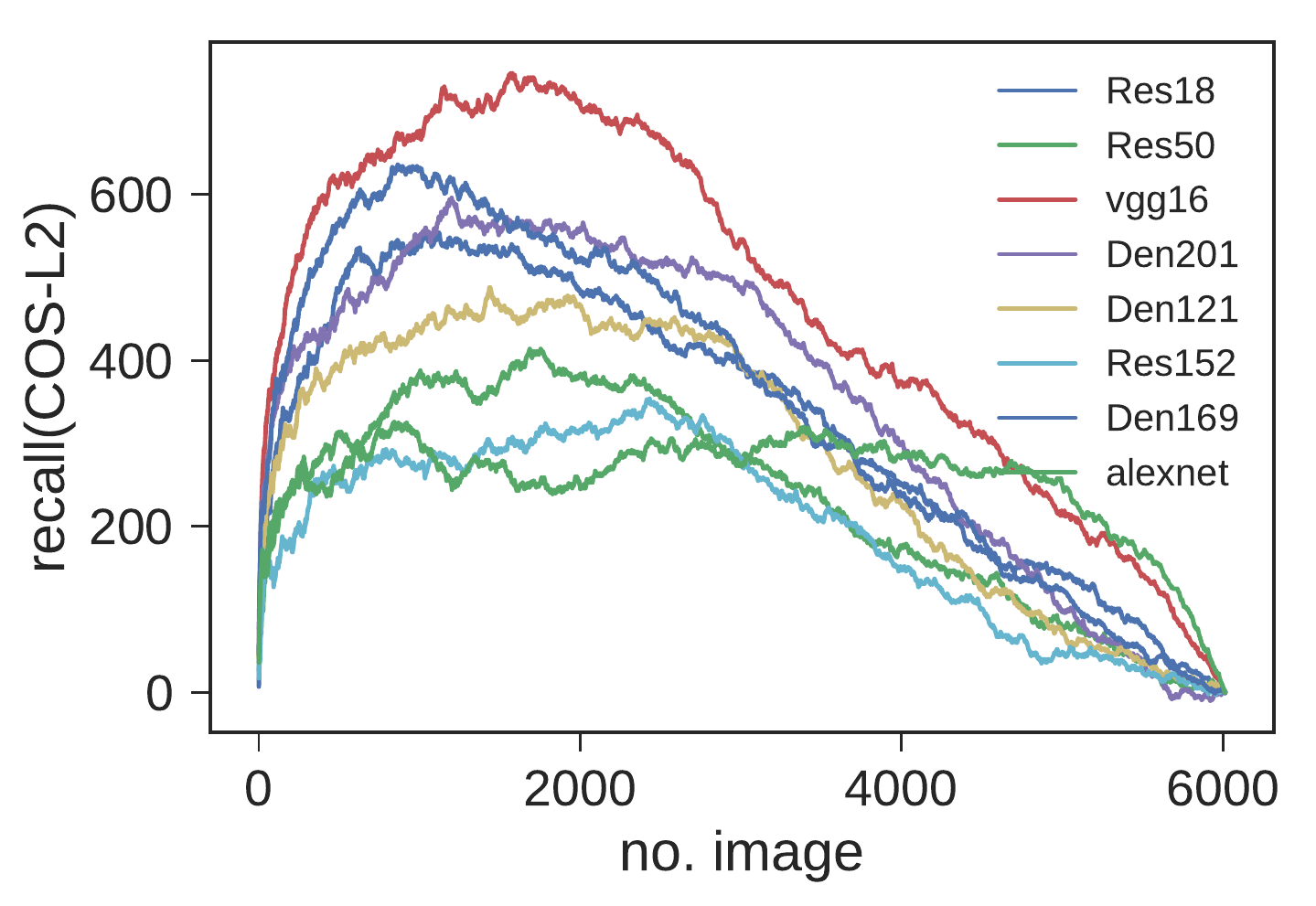} 
\par\end{centering}
}\subfloat[]{\includegraphics[height=0.28\textwidth]{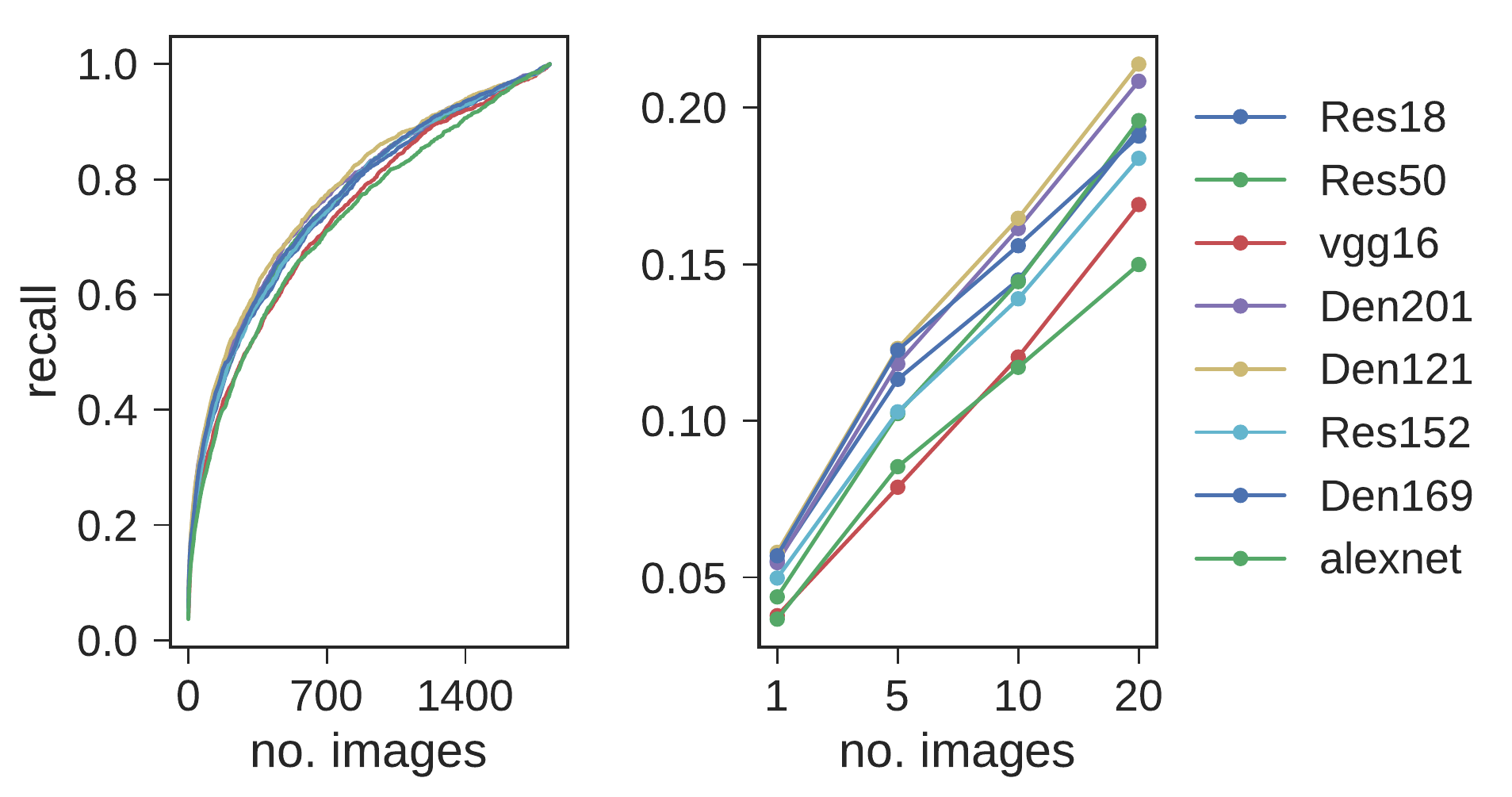}

}
\end{figure}

\textbf{Evaluating Generic Features}: as a first step, we evaluate
which metric (Euclidean vs. cosine) better matches the pairings in
TLL. We noted that the recall for a given number of candidates using
the cosine distance is always higher compared to that of the Euclidean
distance. This can be seen in Figure \ref{fig:Retrieval-Graph} (a).
We calculated recall for each of the nets as a function of the number
of retrieved candidates. The figure shows the difference for each
$k$ between the recall for the cosine vs. Euclidean distances. The
cosine distance has a clear advantage here, hence we choose to use
it for all subsequent experiments (except for the case of facial features).

\textbf{Near duplicates}: visualizing some of the returned nearest
neighbors revealed that there are duplicate (or near duplicate images)
within the $\mathcal{L}$ and $\mathcal{R}$ image sets. As this could
cause an ambiguity and hinder retrieval scores, we removed all pairs
where either the left or the right image was part of the duplicate.
We did this for both generic features and face-based features. For
generic ones, this corresponds to a cosine distance of $\ge0.15$
(using Densenet121); virtually all images below a distance of $0.1$
were near-duplicate, so we set the threshold conservatively to avoid
accidental duplicates. For faces we set the threshold to 0.5. We also
removed duplicates across pairs, meaning that if $L_{i}$ and $R_{j}$
were found to be near-duplicates then we removed them, as an identical
copy $R_{j}$ of $L_{i}$ may be a better match for it than $R_{i}$.
Removing all such duplicates leaves us with a subset we name $TLL_{d}$,
containing 1828 valid image pairs. The results of Table \ref{tab:Retrieval-Table}
and \ref{tab:associative-recall}(a) are calculated based on this
dataset. This does not, however, reduce the importance of the full
dataset of 6016 images as it still contains many interesting and useful
image pairs to learn from. The reduction of the dataset size is only
done for evaluation purposes.

\textbf{Faces: }many images in the dataset contain faces, as indicated
by Figure \ref{fig:Probability-of-agreement} (c). In fact, the figure
represents an underestimation of the number of faces as some faces
we not detected. Such images seem qualitatively different from the
ones containing faces, in that the similarities are more about global
shape, texture, or face-like properties, though there are no actual
faces in them in the strict sense. Hence, we create another partition
of the data without any detected faces, and without the duplicate
images according to the generic feature criteria. This subset, $TLL_{obj}$,
contains 1622 images. Both $TLL_{d}$ and $TLL_{obj}$ are used in
Section \ref{subsec:Human-Experiments} where we report additional
results of human experiments. 

\begin{center}

\begin{table}
\begin{centering}

\begin{tabular}{lrrrrrr}
\toprule 
 & R@1  & R@5  & R@10  & R@2  & R@50  & R@100 \tabularnewline
\midrule 
AlexNet  & 3.67  & 9.19  & 12.09  & 15.37  & 22.59  & 30.63 \tabularnewline
vgg16  & 3.77  & 8.97  & 12.58  & 16.90  & 24.02  & 32.39 \tabularnewline
Res50  & 4.38  & 11.43  & 15.04  & 19.91  & 28.77  & 36.71 \tabularnewline
Res152  & 4.98  & 11.16  & 14.61  & 18.82  & 26.20  & 35.61 \tabularnewline
Den201  & 5.47  & 12.91  & 16.63  & 21.44  & 30.47  & 38.18 \tabularnewline
Res18  & 5.53  & 12.14  & 15.10  & 19.47  & 28.06  & 35.61 \tabularnewline
Den169  & 5.69  & 13.07  & 16.19  & 19.31  & 28.67  & 37.53 \tabularnewline
Den121  & 5.80  & 13.84  & 16.90  & 21.94  & 29.92  & 38.89 \tabularnewline
\bottomrule
\end{tabular}
\par\end{centering}
\caption{\label{tab:Retrieval-Table}Retrieval performance (percentage retrieved
after varying number of candidates) by various learned representations
in the TLL dataset.}
\end{table}
\par\end{center}
Next, we evaluate the retrieval performance as a function of the number
of returned image candidates. This can be seen graphically in Figure
\ref{fig:Retrieval-Graph} (b). The left sub-figure shows the recall
for the entire dataset and the right sub-figure shows it for the first,
5th, 10th and 20th returned candidates. Table \ref{tab:Retrieval-Table}
shows these values numerically. For face features the retrieval accuracy
using one retrieve item was slightly better than the generic features,
reaching 6.1\%. Using generic features extracted on faces performed
quite poorly, at 2.6\%. Evidently, none of the networks we tested
performed well on this benchmark. Such a direct comparison is problematic
for several reasons. Next, we attempt to ease the retrieval task for
the machine-based features.

\textbf{Simulating Associative Recall \label{par:Simulating-Associative-Recall}}:
As mentioned in Sec. \ref{par:Direct-Comparison-vs.}, directly comparing
to all images in the dataset is perhaps unfair to the machine-learning
test. Arguably, a human recalling an image first narrows down the
search given the query image, so only images with relevant features
are retrieved from memory. Though we do not speculate about how this
may be done, we can test how retrieval improves if such a process
were available. To do so, we sample for each left image $L_{i}$ a
random set $R(L_{i})$ of size $m$ which includes the correct right
image $R_{i}$ and an additional $m-1$ images. This simulates a state
where viewing the image $L_{i}$ elicited a recollection of $m$ candidates
(including the correct one) from which the final selection can be
made. We do this for varying sizes of a recollection set $m\in\{1-5,10,20,50,100\}$,
with 10 repetitions each. Table \ref{tab:associative-recall} (a)
summarizes the mean performance obtained here. Although these are
almost ``perfect'' conditions, the retrieval accuracy falls to less
than 50\% if we use as little as ten examples as the test set. The
variance (not shown) was close to 0 in all conditions.

\begin{figure}
\caption{\label{fig:Automatic-retrieval-errors:}Automatic retrieval errors:
using distances between state-of-the-art deep-learned representations
often does not do well in reproducing human similarity judgments.
Each row shows a query image on the left, five retrieved images and
the ground-truth on the right. Perceptual similarity can be attributed
to similarity between cartoonish and real faces (first three rows),
flexible transfer of facial expression (4th row), visually similar
sub-regions (last two rows, hair of person on row 5 resembles spider
legs, hair of person on last row resembles waves). Though the images
and the retrieved ones may be much more similar to each other in a
strict sense, humans still consistently agree on the matched ones
(first, last columns). }

\begin{centering}
\includegraphics[viewport=0bp 0bp 1415bp 245bp,width=0.6\columnwidth]{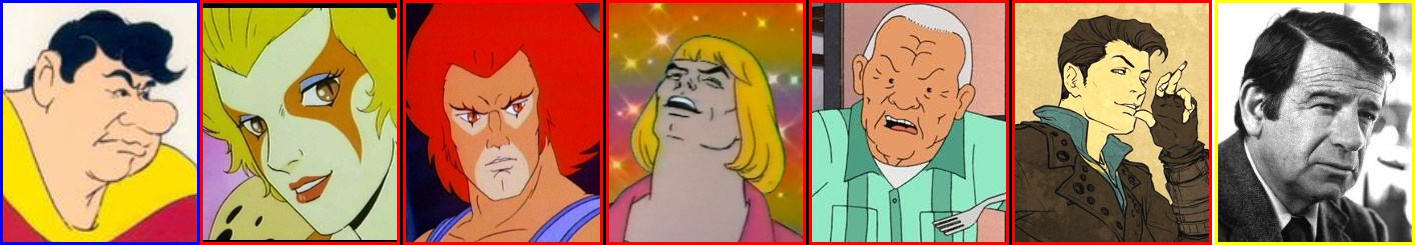} 
\par\end{centering}
\begin{centering}
\includegraphics[viewport=0bp 0bp 1415bp 245bp,width=0.6\columnwidth]{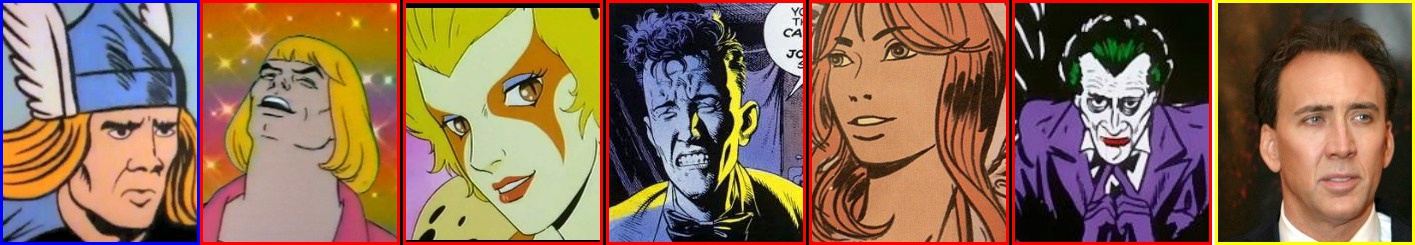} 
\par\end{centering}
\begin{centering}
\includegraphics[viewport=0bp 0bp 1415bp 245bp,width=0.6\columnwidth]{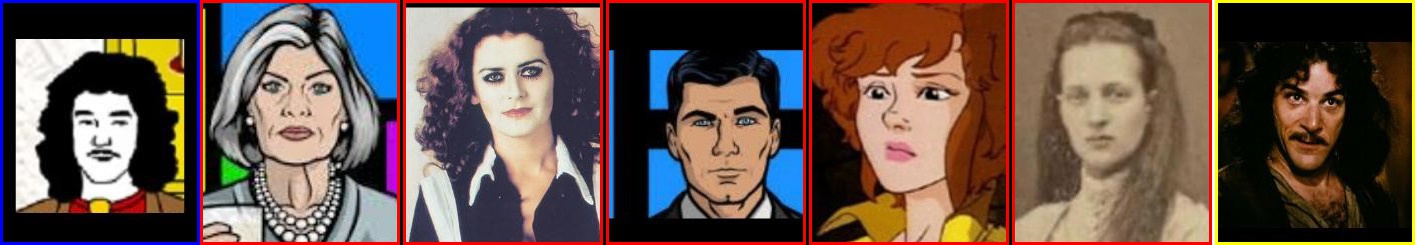} 
\par\end{centering}
\begin{centering}
\includegraphics[viewport=0bp 0bp 1415bp 245bp,width=0.6\columnwidth]{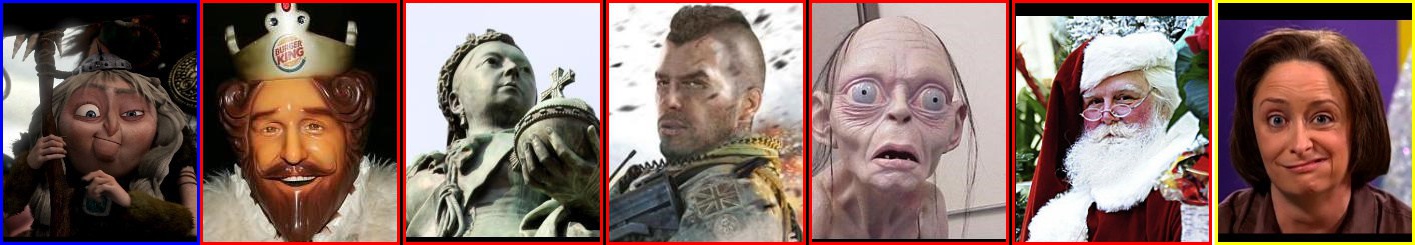} 
\par\end{centering}
\begin{centering}
\includegraphics[viewport=0bp 0bp 1415bp 245bp,width=0.6\columnwidth]{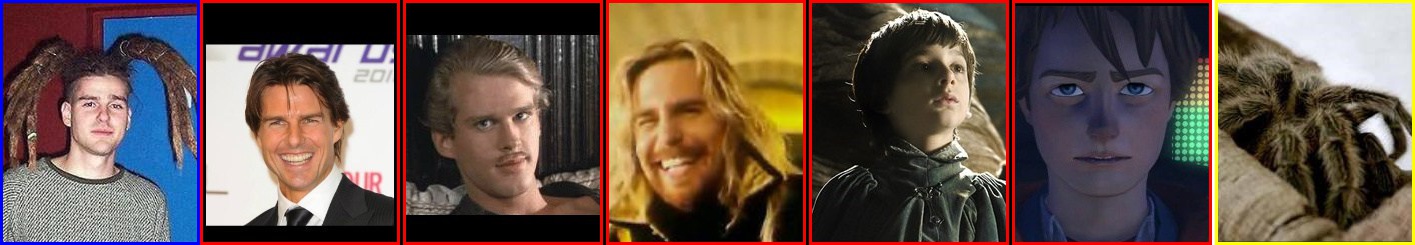} 
\par\end{centering}
\centering{}\includegraphics[viewport=0bp 0bp 1415bp 245bp,width=0.6\columnwidth]{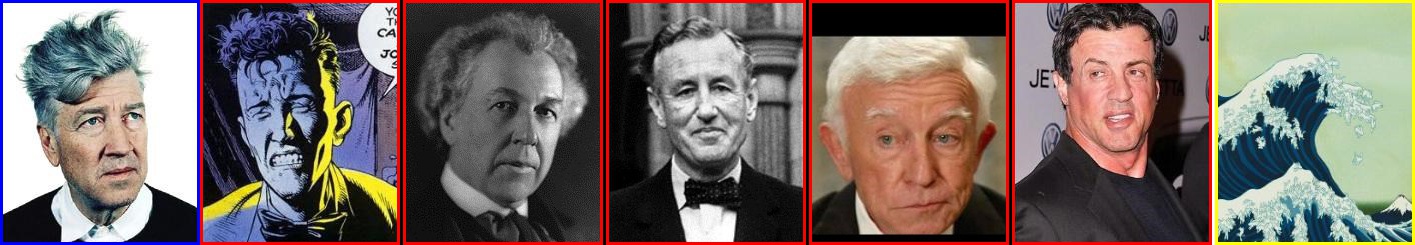} 
\end{figure}

\textbf{Comparing Distances to Votes}: we test whether there is any
consistency between the feature-based distances and the number of
votes assigned by human users. Assuming that a similar number of users
viewed each uploaded image pair, a higher number of votes suggests
higher agreement that the pairing is indeed a valid one. Possibly,
this could also suggest that the images should be easier to match
by automatically extracted features. We calculate the correlation
between number of up-votes and down-votes vs the cosine-distance resulting
from the Densenet121 network. Unfortunately, there seems very little
correlation, with a Pearson coefficient of 0.023 / -0.068 for up/down-votes
respectively. Hence the following experiments do not use the voting
information.

\subsection{Human Experiments \label{subsec:Human-Experiments}}

\begin{table*}
\begin{centering}
\subfloat[]{\begin{centering}
{\small{}{}}%
\begin{tabular}{rr}
\toprule 
$m$  & \% correct \tabularnewline
\midrule 
1  & 100.00 \tabularnewline
2  & 73.35 \tabularnewline
3  & 61.54 \tabularnewline
4  & 54.30 \tabularnewline
5  & 50.49 \tabularnewline
10  & 37.99 \tabularnewline
20  & 27.23 \tabularnewline
50  & 13.37 \tabularnewline
\bottomrule
\end{tabular}
\par\end{centering}
}\subfloat[]{\centering{}%
\begin{tabular}{|c|cc||cccc}
\hline 
 & \multicolumn{2}{c||}{$TLL_{obj}$} & \multicolumn{4}{c}{$TLL_{d}$}\tabularnewline
\hline 
 & random$\dagger$  & generic  & random  & face  & face-generic  & generic\tabularnewline
\hline 
human(lab)  & 83.3  & 70  & 82.5  & 63.3  & 64.5  & 83.3\tabularnewline
human(AMT)  & 84  & 68.25  & 90.25  & 59  & 60.5  & 74.5\tabularnewline
machine  & 20  & 20  & 25  & 0  & 0  & 5\tabularnewline
\end{tabular}

}
\par\end{centering}
\caption{(a) \label{tab:associative-recall}Modeling Associative Recall: percentage
of correct matches using conv-net derived features for the TLL dataset
when a random sample of $m$ images including the correct one is used.
For 10 images, the performance is less that 50\%. (b) \label{tab:Human-Experiments}
man-versus-machine image matching accuracy for the perceptual similarity
task. $\dagger$The relatively high accuracy for ``random'' is because
a small subset is selected which contains the correct answer, highly
increasing the chance for correct guessing. }
\end{table*}

We conducted experiments both in-lab and using Amazon Mechanical Turk
(AMT). We chose 120 random pairs of images from the dataset, as follows:
40 pairs were selected $TLL_{obj}$ and 80 from $TLL_{d}$. From each
pair, we displayed the left image to the user, along with 4 additional
selected images and the correct right image. The images were shuffled
in random order. Human subjects were requested to select the most
similar image to the query (left) image. We allocated 20 images to
each sub-experiment. The names of the experiments are \textbf{random},\textbf{
generic},\textbf{ face }and \textbf{face-generic}, indicating the
type of features used to select the subset, if any. For \textbf{random}
we simply chose a subset of 5 images randomly, similarly to what is
described in \ref{par:Simulating-Associative-Recall}. For each of
the others, we ordered the images from the corresponding subset using
each feature type and retained the top-5. If the top-5 images retrieved
did not contain the correct answer, we randomly replaced one of them
with it. A correct answer in this sense is selecting the correct right
image, for the human, and ranking it highest for the machine. In each
experiment, the four images except the correct match are regarded
as \emph{distractors}. Distractors generated using feature similarity
(as opposed to random selection) pose a greater challenge for human
participants, as they tend to resemble, in some sense, the ``correct''
answer. Table \ref{tab:associative-recall} (b) summarizes the overall
accuracy rates. In lab settings (12 participants, ages 28-39) answered
all 120 questions each (labeled human1,human2 in the table). For AMT,
we repeated each experiment 20 times, where an experiment is answering
a single query, making an overall of 2400 experiments. A payment of
5 cents was rewarded for the completion of each experiment. Only ``master''
workers were used in the experiment, for increased reliability. We
next highlight several immediate conclusions from this data.

\textbf{Data Verification:} the first utility of the collected human
data is to validate the consistency of that collected from the website.
Though not quite perfect, there is large consistency between the human
workers on AMT and the users that uploaded the original TLL images.
The performance of the lab-tested humans seems to be higher on average
than the AMT workers, hinting that either the variability in human
answers is rather large or that the AMT results contain some noise.
Indeed, when we count the number of votes given to each of the five
options, we note a trend to select the first option the most, persisting
through options 2-4. The number of times each option was selected
was 627, 522, 465, 395, 391; option 1 selected 30\% more times than
the expected probability. Nevertheless, we see quite a high agreement
rate throughout the table.

\begin{figure}
\caption{\label{fig:Probability-of-agreement} (a) Probability of agreement
between human users on the AMT experiment. Humans tend to be highly
consistent in their answers. (b) user agreement ratio vs. correct
matching with TLL. (c) Distribution of number of detected faces and
agreement on detected faces between left-right image pairs. }

\centering{}\subfloat[]{\begin{centering}
\includegraphics[height=0.21\columnwidth]{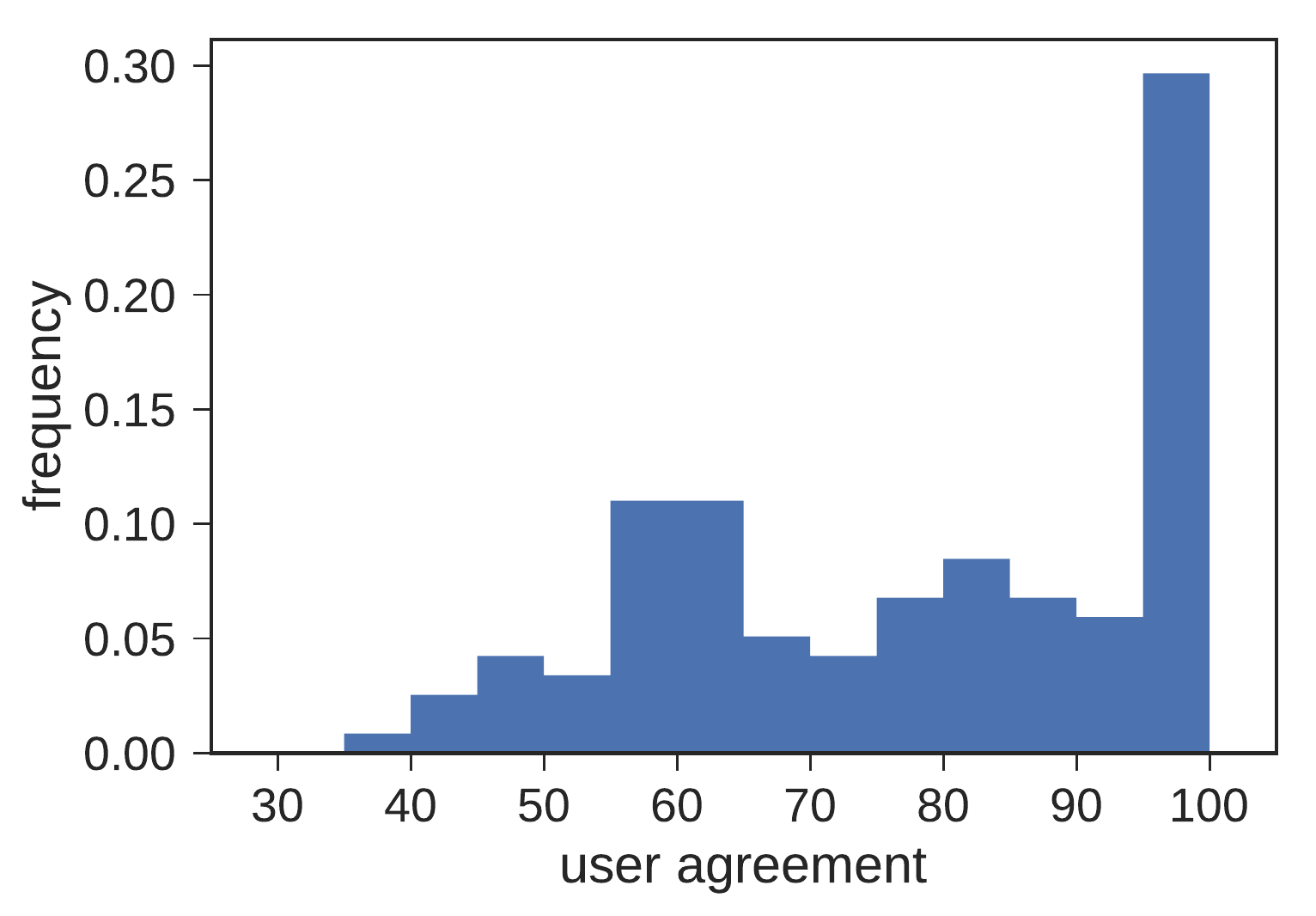} 
\par\end{centering}
}\subfloat[]{\begin{centering}
\includegraphics[height=0.21\columnwidth]{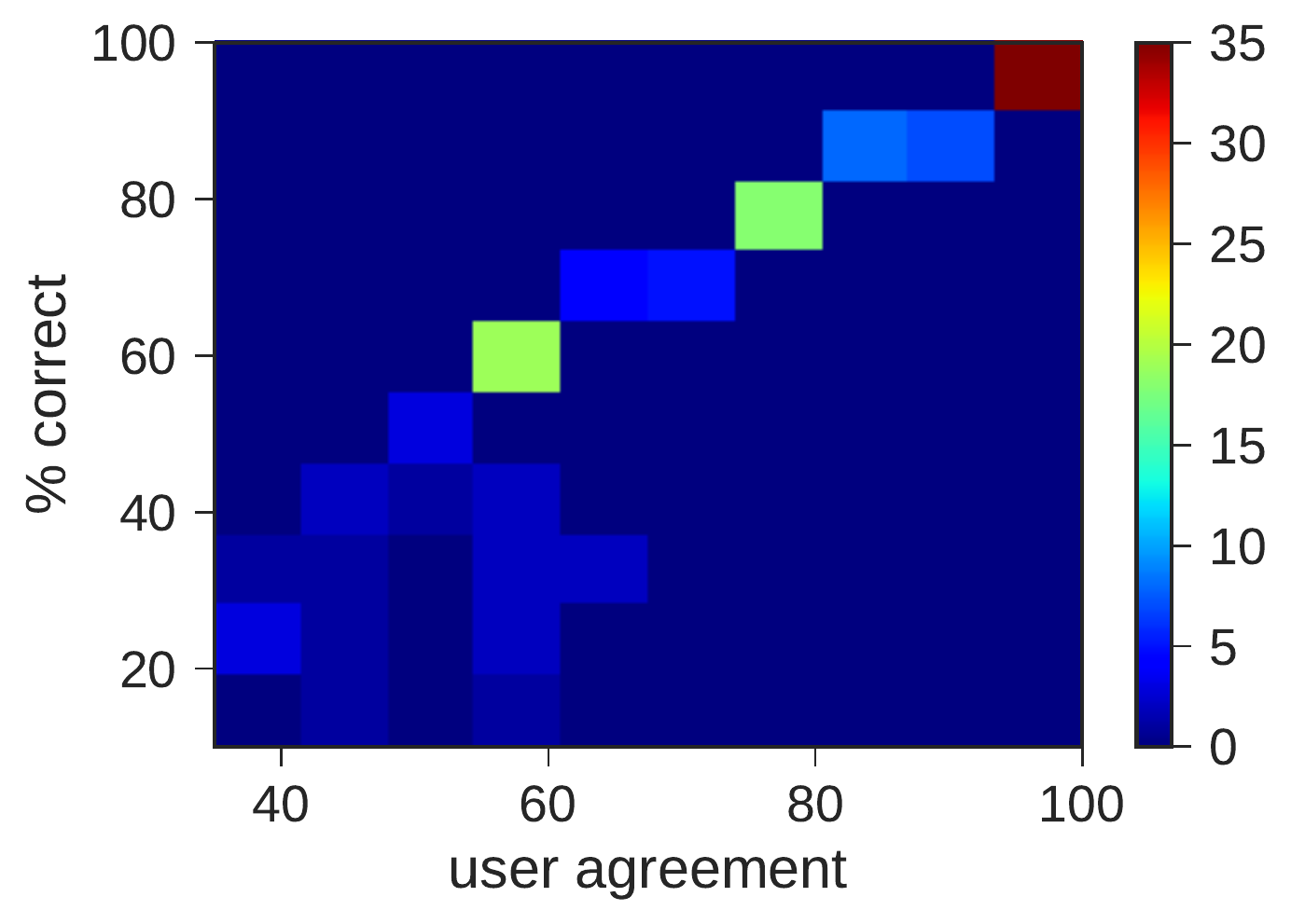} 
\par\end{centering}
}\subfloat[]{\begin{centering}
\includegraphics[height=0.21\columnwidth]{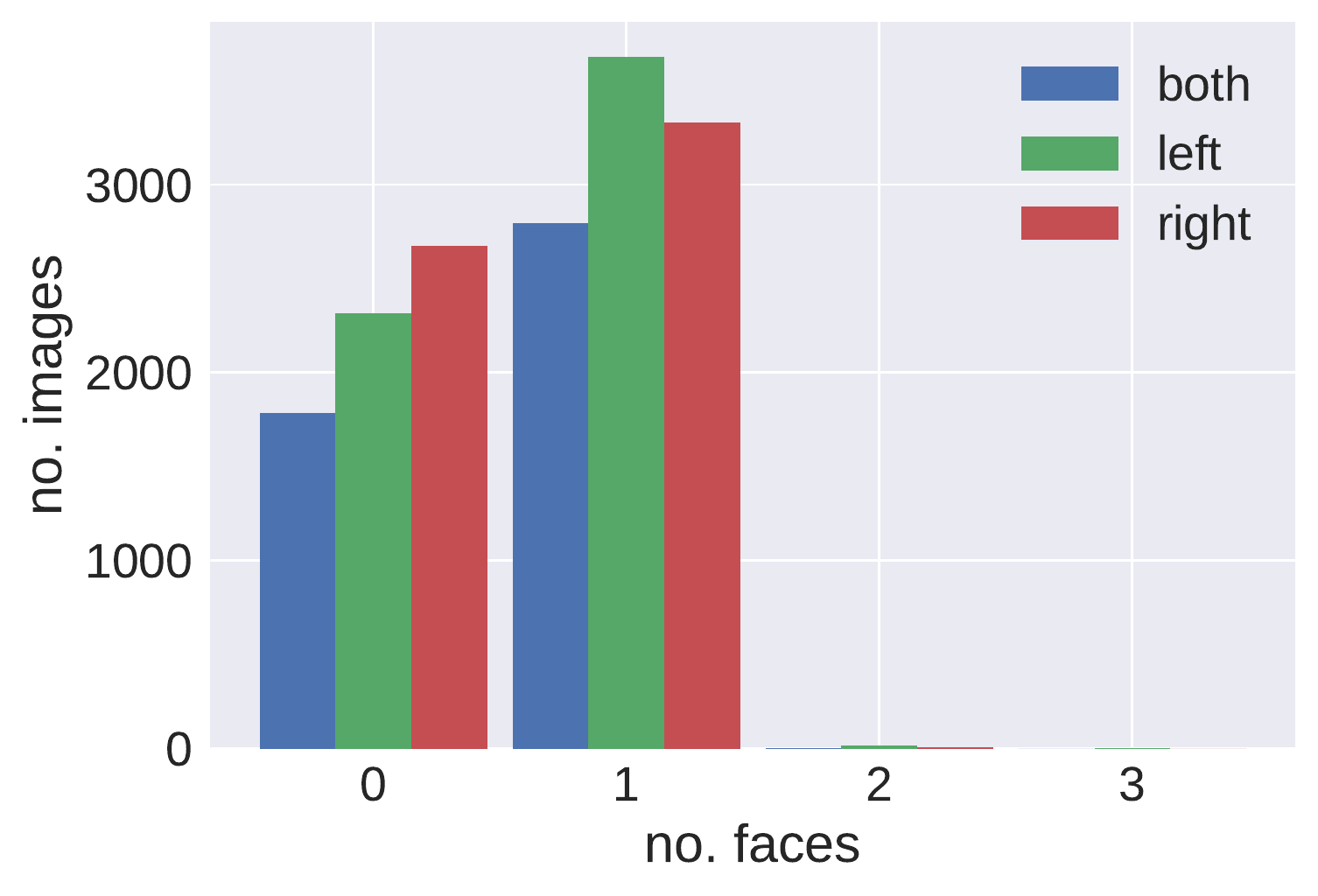} 
\par\end{centering}
}
\end{figure}

\textbf{Human vs Machine Performance:} the average human performance
is generally lower when distractors are selected non-randomly, as
expected. This is especially true for face images, where deep-learned
features are used to select the distractor set; here AMT humans achieve
around 60\% agreement with the TLL dataset. This is not very surprising,
as deep-learned face representations have already been reported to
surpass human performance several years ago \cite{lu2015surpassing}.
This may suggest that for faces, distractor images brought by the
automatic retrieval seemed like better candidates to the humans than
the original matches. The very low consistency of the machine retrieval
with humans is consistent with what is reported in table \ref{tab:Retrieval-Table};
the less than 6\% performance rates translated to 0, in this specific
sample of twenty examples for each test case. The relatively high
performance in the ``random'' cases is due to selection of random
distractors which were likely no closer in feature-space than the
nearest neighbors of the query, hence resulting in seemingly high
performance. We further show the consistency among human users by
counting the number of agreements on answers. We count for each query
the frequency of each answer and test how many times humans agreed
between themselves. In 87\% of the cases, the majority of users (at
least 11 out of 20) agreed on the answer. In fact, the most frequent
event, occurring 30\% of the time, was a total agreement - 20 out
of 20 identical answers. Moreover, the Pearson correlation coefficient
between user agreement and a correct matching to TLL was 0.94. The
plot of agreement frequencies is shown in Figure \ref{fig:Probability-of-agreement}
(a). This large agreement is not in contradiction to the lower rates
of success in reproducing the TLL results, because the TLL dataset
was generated by a different process of unconstrained recollection,
rather than forced choice as in our experiments. Figure \ref{fig:Probability-of-agreement}
(b) shows the relation between user agreement ratios and the distribution
of correctly answered images.

\begin{figure}
\caption{\label{fig:Sample-queries-with}Sample queries with varying user agreement.
Each row shows on the left column a query image and 5 images from
which to select a match. Some queries are very much agreed upon and
on some the answers are evenly distributed. We show two rows of the
first case, and two of the second. We encourage the reader to guess
which images were of each kind. }

\begin{centering}
\includegraphics[width=0.65\textwidth]{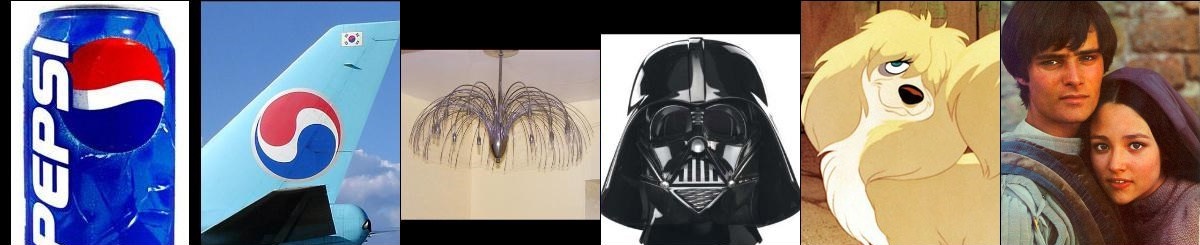}
\par\end{centering}
\begin{centering}
\includegraphics[width=0.65\textwidth]{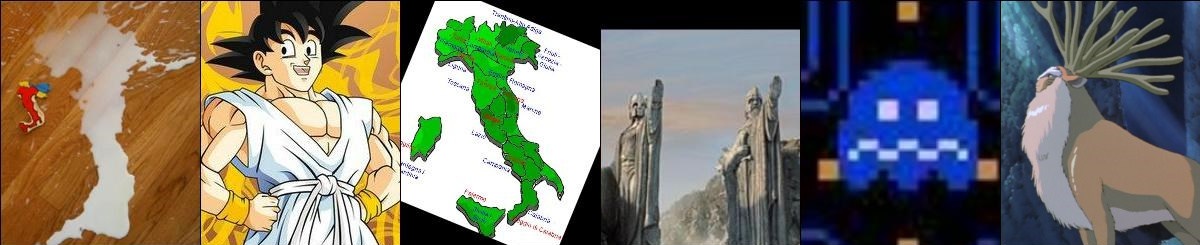}
\par\end{centering}
\begin{centering}
\includegraphics[width=0.65\textwidth]{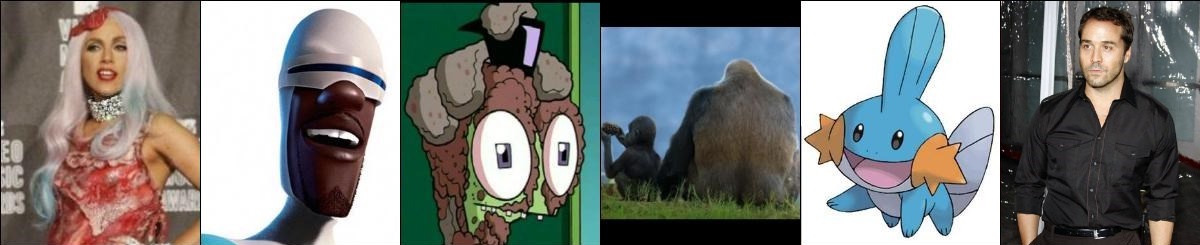}
\par\end{centering}
\centering{}\includegraphics[width=0.65\textwidth]{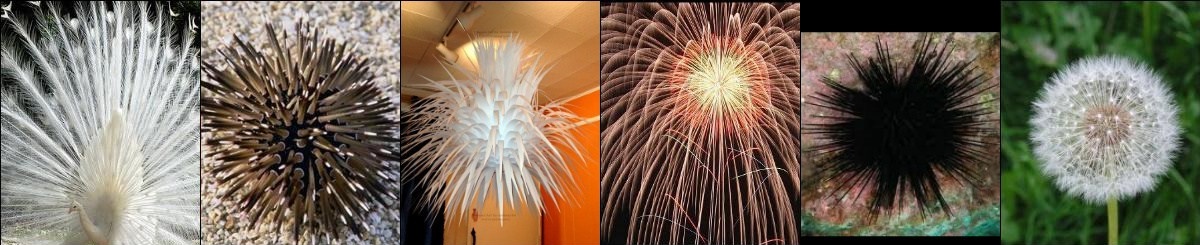} 
\end{figure}

Finally, Figure \ref{fig:Sample-queries-with} shows four queries
from the dataset, in the form of one query image (left column) and
five candidates (remainder columns). Two of the rows shows cases where
there was a perfect human agreement and two show cases where the answers
were almost uniformly spread over the candidates. It is not difficult
to guess which rows represent each case.

\section{Discussion}

We have looked into a high-level task of human vision: perceptual
judgment of image similarity. The new TLL dataset offers a glimpse
into images which are matched by human beings in the "wild", in a
less controlled fashion, but arguably one that sheds a different light on various
factors compared previous work in this area. Most works in image retrieval
deal with near-duplicate images, or images which mostly depict the
same type of concept. We explored the ability of existing state-of-the-art
deep-learned features to reproduce the matchings in the dataset. Though
one would predict this to produce a reasonable baseline, neither features
resulting from object classification networks and ones tailored for
face verification seem to be able to remotely reproduce the matchings
between the image pairs. We verified this using additional human experiments,
both in-lab and using Amazon Mechanical Turk. Tough the collected
data from AMT was not cleaned and clearly showed signs of existence
of biases, the statistics still clearly show that humans are quite
consistent in choosing image pairs, even when faced with a fair amount
of distractors. Emulating easier scenarios for machines (for example,
Table \ref{tab:associative-recall} (a)) yielded improved results,
but ones which are still very far from reproducing the consistency
observed among humans.
\begin{center}
\begin{figure*}
\caption{\label{fig:more-examples}Additional examples. Perceived image similarities
can be abstract/symbolic: cats $\leftrightarrow$ guards, doorway
$\leftrightarrow$ mountain passageway \emph{(a)}, low-level (colors,
\emph{(d,e,f)}, 2D shape \emph{(b,c,e,g)}, 3D-shape \emph{(e)}, related
to well-known iconic images from pop-culture \emph{(b,e,f,h)} art
\emph{(c)} or pose-transfer across very different objects/domains
\emph{(b,c,d)} }

\begin{centering}
\begin{minipage}[t]{0.24\textwidth}%
\includegraphics[viewport=0bp 23bp 401bp 271bp,clip,scale=0.25]{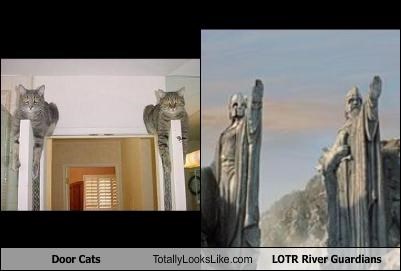}\\(a)%
\end{minipage}\,%
\begin{minipage}[t]{0.24\textwidth}%
\includegraphics[viewport=0bp 23bp 401bp 271bp,clip,scale=0.25]{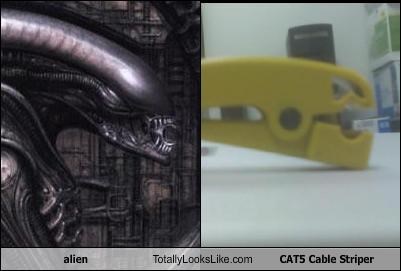}
(b)%
\end{minipage}\,%
\begin{minipage}[t]{0.24\textwidth}%
\includegraphics[viewport=0bp 23bp 401bp 271bp,clip,scale=0.25]{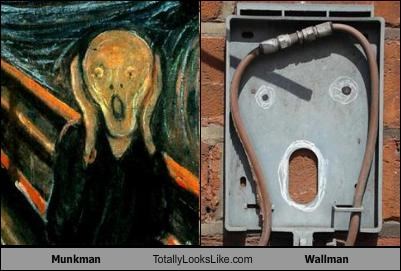}\\(c)%
\end{minipage}\,%
\begin{minipage}[t]{0.24\textwidth}%
\includegraphics[viewport=0bp 23bp 401bp 271bp,clip,scale=0.25]{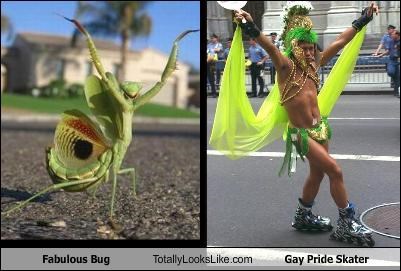}\\(d)%
\end{minipage}
\par\end{centering}
\centering{}%
\begin{minipage}[t]{0.24\textwidth}%
\includegraphics[viewport=0bp 23bp 401bp 271bp,clip,scale=0.25]{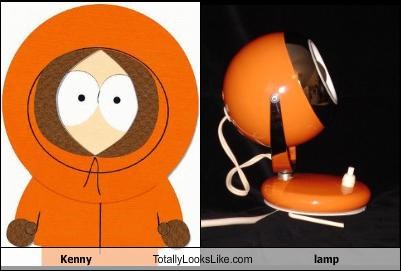}\\(e)%
\end{minipage}\,%
\begin{minipage}[t]{0.24\textwidth}%
\includegraphics[viewport=0bp 23bp 401bp 271bp,clip,scale=0.25]{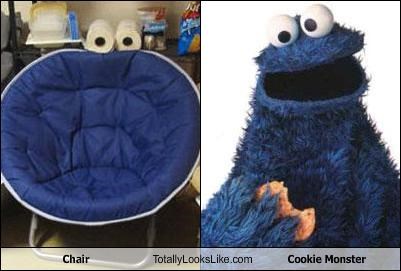}\\(f)%
\end{minipage}\,%
\begin{minipage}[t]{0.24\textwidth}%
\includegraphics[viewport=0bp 23bp 401bp 271bp,clip,scale=0.25]{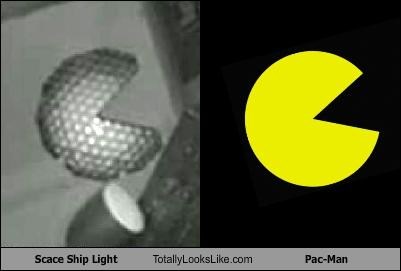}\\(g)%
\end{minipage}\,%
\begin{minipage}[t]{0.24\textwidth}%
\includegraphics[viewport=0bp 23bp 401bp 271bp,clip,scale=0.25]{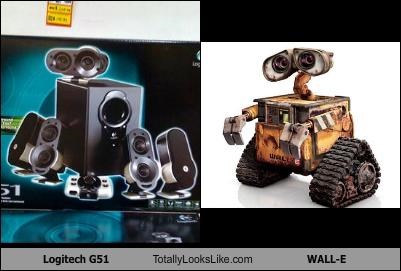}\\(h)%
\end{minipage}
\end{figure*}
\par\end{center}

One could argue that fine-tuning the machine learned representation
with a subset of images in this dataset will reduce the observed gap.
However, we believe that sufficiently generic visual features should
be able to reproduce the same similarity measurements without being
explicitly trained to do so, just as humans do. Moreover, the set
of various features employed by humans is likely rather large; previous
attempts to reproduce human similarity measurements resulted in datasets
much larger than the proposed one, though they were narrower in scope
in terms of image variability (for example \cite{pramod2016computational}).
This raises the question, how many images will an automatic method
require to reproduce this rich set of similarities demonstrated by
humans?

We do not expect strong retrieval systems to reproduce the matchings
in TLL. On the contrary, a cartoon figure should not be automatically
associated with the face of Nicolas Cage \ref{fig:Automatic-retrieval-errors:}
(2nd row), this would likely constitute a retrieval error in normal
conditions and lead to additional unexpected ones. However, we do
expect a high-level representation to report that of all the
images in that row, the most similar one is indeed that
of the said actor. Humans can easily point to the facial features
in which the cartoon and the natural face image bear resemblance.
In fact, we believe that for similarity judgments to be consistent
with those of humans (note there is no ``correct'' or ``incorrect''),
they should be multi-modal and conditioned on \emph{both} images.
Relevant factors include (1) facial features (2)
facial expressions (3rd row in Figure \ref{fig:Automatic-retrieval-errors:}),
requiring a robust comparison between facial expressions in different
modalities (3) texture or structure of part of
the image (last row, person's hair). The factors are not fixed or weighted equally in each case. Additional
factors involve comparison between different objects or familiarity with iconic images or characters as depicted in Figure \ref{fig:more-examples}.

As the importance of factors changes as a function of the image-pair,
we suggest that the comparison will be akin to visual-question-answering
(VQA) , in the form ``\emph{why should image A be regarded as similar
/ dissimilar to image B?}''. Just as VQA models on single images
benefit from attention models \cite{xu2015show}, we suggest that
asking a question that requires extracting relevant information from
two different images will give rise to attention being applied to
both. Information extracted from one image (such as the presence of
a face, waves, an unusual facial expression, or spider-legs in Figure
\ref{fig:Automatic-retrieval-errors:}) is necessary to produce a
basis for comparison and feature extraction from the other. We leave
further development of this direction to future work.

\bibliographystyle{splncs04}
\bibliography{eccv2018submission}

\begin{thebibliography}{10}
\providecommand{\url}[1]{\texttt{#1}}
\providecommand{\urlprefix}{URL }
\providecommand{\doi}[1]{https://doi.org/#1}

\bibitem{antol2015vqa}
Antol, S., Agrawal, A., Lu, J., Mitchell, M., Batra, D., {Lawrence Zitnick},
  C., Parikh, D.: {Vqa: Visual question answering}. In: {Proceedings of the
  IEEE International Conference on Computer Vision}. pp. 2425--2433 (2015)

\bibitem{battleday2017modeling}
Battleday, R.M., Peterson, J.C., Griffiths, T.L.: {Modeling human
  categorization of natural images using deep feature representations}. arXiv
  preprint arXiv:1711.04855  (2017)

\bibitem{brady2008visual}
Brady, T.F., Konkle, T., Alvarez, G.A., Oliva, A.: {Visual long-term memory has
  a massive storage capacity for object details}. Proceedings of the National
  Academy of Sciences  \textbf{105}(38),  14325--14329 (2008)

\bibitem{chandrasekaran2016we}
Chandrasekaran, A., Vijayakumar, A.K., Antol, S., Bansal, M., Batra, D.,
  {Lawrence Zitnick}, C., Parikh, D.: {We are humor beings: Understanding and
  predicting visual humor}. In: {Proceedings of the IEEE Conference on Computer
  Vision and Pattern Recognition}. pp. 4603--4612 (2016)

\bibitem{das2017human}
Das, A., Agrawal, H., Zitnick, L., Parikh, D., Batra, D.: {Human attention in
  visual question answering: Do humans and deep networks look at the same
  regions?} Computer Vision and Image Understanding  \textbf{163},  90--100
  (2017)

\bibitem{deza2015understanding}
Deza, A., Parikh, D.: {Understanding image virality}. In: {Proceedings of the
  IEEE conference on computer vision and pattern recognition}. pp. 1818--1826
  (2015)

\bibitem{geirhos2017comparing}
Geirhos, R., Janssen, D.H., Sch{\"u}tt, H.H., Rauber, J., Bethge, M., Wichmann,
  F.A.: {Comparing deep neural networks against humans: object recognition when
  the signal gets weaker}. arXiv preprint arXiv:1706.06969  (2017)

\bibitem{he2016deep}
He, K., Zhang, X., Ren, S., Sun, J.: {Deep residual learning for image
  recognition}. In: {Proceedings of the IEEE conference on computer vision and
  pattern recognition}. pp. 770--778 (2016)

\bibitem{huang2016densely}
Huang, G., Liu, Z., Weinberger, K.Q., van~der Maaten, L.: {Densely connected
  convolutional networks}. arXiv preprint arXiv:1608.06993  (2016)

\bibitem{10.3389/fpsyg.2017.01726}
Jozwik, K.M., Kriegeskorte, N., Storrs, K.R., Mur, M.: {Deep Convolutional
  Neural Networks Outperform Feature-Based But Not Categorical Models in
  Explaining Object Similarity Judgments}. Frontiers in Psychology  \textbf{8},
  ~1726 (2017). \doi{10.3389/fpsyg.2017.01726},
  \url{https://www.frontiersin.org/article/10.3389/fpsyg.2017.01726}

\bibitem{ICCV15_Khosla}
Khosla, A., Raju, A.S., Torralba, A., Oliva, A.: {Understanding and Predicting
  Image Memorability at a Large Scale}. In: {International Conference on
  Computer Vision (ICCV)} (2015)

\bibitem{konkle2010scene}
Konkle, T., Brady, T.F., Alvarez, G.A., Oliva, A.: {Scene memory is more
  detailed than you think: The role of categories in visual long-term memory}.
  Psychological Science  \textbf{21}(11),  1551--1556 (2010)

\bibitem{krizhevsky2012imagenet}
Krizhevsky, A., Sutskever, I., Hinton, G.E.: {Imagenet classification with deep
  convolutional neural networks}. In: {Advances in neural information
  processing systems}. pp. 1097--1105 (2012)

\bibitem{liu2017survey}
Liu, W., Wang, Z., Liu, X., Zeng, N., Liu, Y., Alsaadi, F.E.: {A survey of deep
  neural network architectures and their applications}. Neurocomputing
  \textbf{234},  11--26 (2017)

\bibitem{lu2015surpassing}
Lu, C., Tang, X.: {Surpassing Human-Level Face Verification Performance on LFW
  with GaussianFace.} In: {AAAI}. pp. 3811--3819 (2015)

\bibitem{peterson2016adapting}
Peterson, J.C., Abbott, J.T., Griffiths, T.L.: {Adapting deep network features
  to capture psychological representations}. arXiv preprint arXiv:1608.02164
  (2016)

\bibitem{pramod2016computational}
Pramod, R., Arun, S.: {Do computational models differ systematically from human
  object perception?} In: {Proceedings of the IEEE Conference on Computer
  Vision and Pattern Recognition}. pp. 1601--1609 (2016)

\bibitem{russakovsky2015imagenet}
Russakovsky, O., Deng, J., Su, H., Krause, J., Satheesh, S., Ma, S., Huang, Z.,
  Karpathy, A., Khosla, A., Bernstein, M., et~al.: {Imagenet large scale visual
  recognition challenge}. International Journal of Computer Vision
  \textbf{115}(3),  211--252 (2015)

\bibitem{schmidhuber2015deep}
Schmidhuber, J.: {Deep learning in neural networks: An overview}. Neural
  networks  \textbf{61},  85--117 (2015)

\bibitem{sharif2014cnn}
{Sharif Razavian}, A., Azizpour, H., Sullivan, J., Carlsson, S.: {CNN features
  off-the-shelf: an astounding baseline for recognition}. In: {Proceedings of
  the IEEE Conference on Computer Vision and Pattern Recognition Workshops}.
  pp. 806--813 (2014)

\bibitem{simonyan2014very}
Simonyan, K., Zisserman, A.: {Very deep convolutional networks for large-scale
  image recognition}. arXiv preprint arXiv:1409.1556  (2014)

\bibitem{workman2016quantifying}
Workman, S., Souvenir, R., Jacobs, N.: {Quantifying and Predicting Image
  Scenicness}. arXiv preprint arXiv:1612.03142  (2016)

\bibitem{xu2015show}
Xu, K., Ba, J., Kiros, R., Cho, K., Courville, A., Salakhudinov, R., Zemel, R.,
  Bengio, Y.: {Show, attend and tell: Neural image caption generation with
  visual attention}. In: {International Conference on Machine Learning}. pp.
  2048--2057 (2015)

\bibitem{zhang2018unreasonable}
Zhang, R., Isola, P., Efros, A.A., Shechtman, E., Wang, O.: {The Unreasonable
  Effectiveness of Deep Features as a Perceptual Metric}. arXiv preprint
  arXiv:1801.03924  (2018)

\bibitem{zhou2017landscape}
Zhou, P., Feng, J.: {The Landscape of Deep Learning Algorithms}. arXiv preprint
  arXiv:1705.07038  (2017)

\bibitem{zhou2017recent}
Zhou, W., Li, H., Tian, Q.: {Recent Advance in Content-based Image Retrieval: A
  Literature Survey}. arXiv preprint arXiv:1706.06064  (2017)

\end{thebibliography}

\end{document}